\documentclass[conference]{IEEEtran}
\IEEEoverridecommandlockouts
\usepackage[top=1in, bottom=0.75in, left=0.75in, right=0.75in]{geometry}
\usepackage{amsmath,amssymb}
\usepackage{algorithm}
\usepackage{subcaption}
\usepackage{booktabs}
\usepackage{multirow}
\usepackage{xcolor} 
\usepackage{soul}

\usepackage{algorithmic}
    \usepackage{graphicx} 

\usepackage{fancyhdr}
\title{Interactive Double Deep Q-network: Integrating Human
Interventions and Evaluative Predictions in Reinforcement Learning of Autonomous Driving}

\usepackage{orcidlink}  

\author{
\IEEEauthorblockN{
Alkis Sygkounas\textsuperscript{1}\orcidlink{0009-0007-4357-9533},  
Ioannis Athanasiadis\textsuperscript{2}\orcidlink{0000-0002-5213-6757},  
Andreas Persson\textsuperscript{1}\orcidlink{0000-0001-7649-9109},  
Michael Felsberg\textsuperscript{2}\orcidlink{0000-0002-6096-3648},  
Amy Loutfi\textsuperscript{1}\orcidlink{0000-0002-3122-693X}
}
\IEEEauthorblockA{\textsuperscript{1} Center for Applied Autonomous Sensor Systems (AASS), Örebro University, Sweden}  
\IEEEauthorblockA{\textsuperscript{2} Computer Vision Laboratory, Linköping University, Sweden}  
\IEEEauthorblockA{Email: \{alkis.sygkounas, andreas.persson, amy.loutfi\}@oru.se,  
\{ioannis.athanasiadis, michael.felsberg\}@liu.se}
}

\begin{document}

\maketitle

\begin{abstract}
Integrating human expertise with machine learning is crucial for applications demanding high accuracy and safety, such as autonomous driving. This study introduces Interactive Double Deep Q-network (iDDQN), a Human-in-the-Loop (HITL) approach that enhances Reinforcement Learning (RL) by merging human insights directly into the RL training process, improving model performance. 
Our proposed iDDQN method modifies the Q-value update equation to integrate human and agent actions, establishing a collaborative approach for policy development. Additionally, we present an offline evaluative framework that simulates the agent’s trajectory as if no human intervention to assess the effectiveness of human interventions. Empirical results in simulated autonomous driving scenarios demonstrate that iDDQN outperforms established approaches, including Behavioral Cloning (BC), HG-DAgger, Deep Q-Learning from Demonstrations (DQfD), and vanilla DRL in leveraging human expertise for improving performance and adaptability.
\end{abstract}

\section{INTRODUCTION}


Achieving autonomous driving remains a key challenge in developing intelligent vehicles capable of reliably perceiving their environment, making real-time decisions, and executing precise control in dynamic and uncertain environments. Deep Reinforcement Learning (DRL) has demonstrated great potential in autonomous driving \cite{zhao2024survey, ahmed2022smart}, as well as in high-dimensional control problems \cite{mnih2013playing}, reward optimization \cite{hazra2024revolve}, and decision-making and control under dynamic conditions \cite{gorostiza2020deep}. However, a persistent challenge remains in effectively integrating human expertise into DRL models to enhance safety, adaptability, and interpretability.


Human-in-the-Loop (HITL) learning has been introduced as a means to address this issue by incorporating human feedback into the training process, allowing for real-time corrections and guidance in complex, high-risk driving scenarios \cite{wu2023toward, wang2023efficient,arakawa2018dqn, li2022efficient}.
A key approach within HITL systems is Imitation Learning (IL), which seeks to learn driving policies from expert demonstrations. 
Among IL methods, Behavioral Cloning (BC) \cite{torabi2018behavioral} presents the most straightforward approach, using pre-collected expert demonstrations to train agents directly. However, the static nature of BC limits its adaptability, as it cannot incorporate human feedback to correct agent errors during deployment or guide learning during training.

To address these challenges, Deep Q-Learning from Demonstrations (DQfD) \cite{hester2018deep} leverages expert demonstrations for pretraining, similar to Behavioral Cloning (BC), but further fine-tunes the policy using reinforcement learning. This approach retains the benefits of supervised learning while allowing the agent to improve beyond the limitations of static demonstrations through interaction with the environment. Meanwhile, HG-DAgger \cite{kelly2019hg} introduces a more interactive approach by incorporating expert corrections iteratively during training, ensuring that the agent continuously refines its policy based on real-time human interventions. While progress has been made in integrating human input into DRL, a notable gap exists in effectively integrating human-agent collaboration. Existing research has often positioned the human as a supervisor, primarily emphasizing correction over active guidance or providing demonstrations in an offline setting, limiting real-time adaptability. This constraint limits the development of a truly collaborative approach, where human and agent inputs dynamically influence training and decision-making. Moreover, current approaches also lack a mechanism to validate the effectiveness of human interventions compared to agent-only decisions. 


To address these gaps, we propose the Interactive Double Deep Q-Network (iDDQN), an interactive DRL framework based on Clipped Double DQN \cite{fujimoto2018addressing}. iDDQN modifies the Q-value update equation to enable integration of real-time human interventions into the training process. This approach allows human and agent actions to be blended dynamically, fostering a collaborative policy that aligns with human intentions. An application domain where human interventions can significantly enhance decision-making is the dynamic and uncertain domain of autonomous driving. As demonstrated by Wu et al.~\cite{wu2023toward}, real-time human guidance in DRL-based autonomous driving is valuable for enabling agents to adapt to unexpected obstacles and environmental variations. Building upon this insight, we evaluate our proposed iDDQN approach using the AirSim simulator \cite{shah2018airsim}, a high-fidelity simulator that replicates complex driving conditions and facilitates real-time human interventions\footnote{While our primary evaluation focuses on the domain of autonomous driving, the iDDQN approach is designed for broader applicability across any domain that can be formulated as a Markov Decision Process (MDP), extending its relevance to various Human-in-the-Loop (HITL) scenarios.}.
Additionally, we propose an evaluative framework for post-hoc assessment of human interventions. This framework compares human inputs with agent-generated actions by estimating the cumulative rewards over a near-future horizon, providing a rigorous measure of the impact of human contributions.

The key contributions of this work are as follows:
\begin{enumerate}
    \item We propose the Interactive Double Deep Q-Network (iDDQN), a novel DRL method that merges Human-in-the-Loop interventions with agent decision-making, enabling the learning of unified, collaborative policies. 

    \vspace{0.16cm}
    
    \item We further introduce an evaluative framework for qualitative comparison between human interventions and the agent’s potential output by comparing human and agent-generated actions based on estimated cumulative rewards.
    
    \vspace{0.16cm}
    
    

    \item Through extensive evaluations, we show that iDDQN outperforms HITL methods, including BC, DQfD, and HG-DAgger, achieving improved policy refinement, enhanced generalization, and faster convergence. Our results characterize the impact of human feedback on learning efficiency and policy performance, quantifying its benefits across different intervention strategies.

\end{enumerate}

\section{Related Work}



Recent advancements in Reinforcement Learning (RL) have significantly improved algorithmic efficiency and policy quality. The introduction of Deep Q-Learning (DQN) \cite{mnih2013playing} marked a milestone by employing deep neural networks to approximate Q-value functions, enabling RL to handle high-dimensional state-action spaces. Building upon the foundation of DQN, Double DQN \cite{van2016deep} resolved overestimation biases by decoupling action selection from evaluation, improving policy quality across various domains. The Dueling Architecture \cite{wang2016dueling} improved learning by separating state-value and advantage-value functions, enhancing decision-making, particularly in vision-based scenarios. Clipped Double Q-Learning \cite{fujimoto2018addressing} enhanced stability in continuous action spaces, while Prioritized Experience Replay (PER) \cite{schaul2015prioritized} increased sample efficiency by prioritizing transitions with higher learning potential.

Human-in-the-Loop (HITL) methods have emerged as a critical component in enhancing RL for complex and dynamic environments. Behavioral Cloning (BC)~\cite{torabi2018behavioral} represents a foundational imitation learning approach that uses pre-collected expert demonstrations to train models via supervised learning. However, BC lacks adaptability for addressing errors during deployment. More interactive forms of Imitation Learning (IL), such as HG-DAgger~\cite{kelly2019hg}, incorporate expert interventions during training to correct catastrophic mistakes and iteratively refine policies through data aggregation. Similarly, Deep Q-Learning from Demonstrations (DQfD)~\cite{hester2018deep} integrates human demonstrations into RL by pretraining on expert data, then fine-tuning with reinforcement learning while leveraging prioritized experience replay and an imitation loss.

Interactive feedback mechanisms have been explored to bridge these gaps with surveys such as \cite{wu2022survey} highlight the diverse applications of HITL, including active learning \cite{monarch2021human} and real-time feedback \cite{arakawa2018dqn, loftin2014learning, lfd2016}. Recent works, like \cite{liu2022robot}, focus on combining human feedback with simultaneous deployment, showcasing the potential of HITL in real-world settings. Strategies such as HACO \cite{li2022efficient} minimize the reliance on human intervention while ensuring safe agent behavior. Meanwhile, \cite{wu2023toward} demonstrated significant performance improvements by integrating real-time human guidance into DRL agents.

Although these methods have advanced HITL learning, several limitations persist. Some approaches, such as BC and DQfD, rely on static, pre-recorded datasets, limiting their adaptability in dynamic or unforeseen scenarios. Others, like HG-DAgger, incorporate expert interventions during training but lack systematic evaluation mechanisms for understanding the necessity or effectiveness of corrections. Additionally, methods such as \cite{li2022efficient, wu2023toward} integrate real-time feedback but do not dynamically blend human and agent actions or provide robust frameworks for quantifying the impact of human contributions. These gaps highlight the need for a more adaptive and systematic framework that dynamically integrates human interventions and rigorously evaluates their impact on policy performance.



\section{Proposed Method}
The proposed method builds upon established Reinforcement Learning (RL) techniques, particularly Clipped Double Q-Learning~\cite{fujimoto2018addressing}, which improves stability by reducing Q-value overestimation through the use of two target networks. This builds on earlier advances such as Deep Q-Learning (DQN)~\cite{mnih2013playing} and Double DQN~\cite{van2016deep}. Additionally, we leverage the Dueling Architecture~\cite{wang2016dueling} to separate state-value and advantage-value functions, and Prioritized Experience Replay (PER)~\cite{schaul2015prioritized} to improve sample efficiency. These techniques form the foundation of our proposed method, Interactive Double Deep Q-Network (iDDQN), which extends Clipped Double Q-Learning by incorporating human interventions dynamically. For details on the underlying techniques, we refer the reader to Appendix~\ref{appendix_preliminaries}.

    \subsection{Interactive Double Deep Q-Network (iDDQN)}\label{pseud_oalgo} 
 
    During each interaction with the environment, the agent records the following transition data: the current state \(s\), the agent's action \(a_{\text{agent}}\), the reward \(r\), and the next state \(s'\). If a human intervention occurs, the human action \(a_{\text{human}}\) is also recorded; otherwise, \(a_{\text{human}}\) is marked as \(-1\) to indicate no intervention. The presence of human intervention is denoted by \(I(s)\), a binary indicator where \(I(s) = 1\) signifies an intervention:

    \begin{equation}
        a_{\text{sampled}} {=} [1 - I(s)] a_{\text{agent}} + I(s) a_{\text{human}},
    \end{equation}

   For each state \(s\), the framework evaluates the potential outcomes of both human and agent-generated actions using the two Q-networks, \(Q_1\) and \(Q_2\), as introduced in Clipped Double Q-Learning \cite{fujimoto2018addressing}. The Q-values for actions are computed as:
\begin{align}
    Q_{1,\text{human}} &= Q_1(s, a_{\text{human}}; \theta_1), \quad Q_{2,\text{human}} = Q_2(s, a_{\text{human}}; \theta_2) \\
    Q_{1,\text{agent}} &= Q_1(s, a_{\text{agent}}; \theta_1), \quad Q_{2,\text{agent}} = Q_2(s, a_{\text{agent}}; \theta_2)
\end{align}
where \(Q_{i,\text{human}} = 0\) if \(a_{\text{human}} = -1\), indicating no human intervention.

To combine the Q-values from both human and agent actions, the method employs a hyperparameter \( \lambda_h \), the human weight factor, to prioritize the impact of the human actions. Specifically, \( \lambda_h \in [0,1] \), where \(\lambda_h = 0\) corresponds to pure reinforcement learning, and \(\lambda_h = 1\) corresponds to fully human-driven decisions. The composite Q-value \(Q_{\text{combined}}\) is derived by weighting the Q-values from both human and agent actions:

\begin{align}
    Q_{\text{combined}} = &\ \lambda_h \min(Q_{1,\text{human}}, Q_{2,\text{human}}) \nonumber \\
    &+ (1 - \lambda_h) \min(Q_{1,\text{agent}}, Q_{2,\text{agent}}),
    \label{eq:combined_q}
\end{align}

We also consider a decaying schedule for the human weight factor $\lambda_h$. 
In this setup, $\lambda_h$ starts at 1.0 and gradually decreases to 0.0 as 
training progresses, progressively shifting decision-making from human-guided 
choices to the autonomous agent.

The target Q-value is computed using the minimum Q-value across the two networks to address overestimation bias:

\begin{equation}
 Q_\text{target} = r {+} \gamma \min_{i=1,2} Q_i(s', \text{argmax}_{a'} Q(s', a'; \theta); \theta^-_i) (1 - \mathrm{done})
\end{equation}

where \(s'\) is the next state, \(\theta^-_i\) are the parameters of the target networks, and \(\textit{done}\) indicates episode termination.

Finally, the Temporal Difference (TD) error is calculated as:
\begin{equation}
    \text{TD}_{\text{error}} = Q_{\text{target}} - Q_{\text{combined}}.
\end{equation}

The iDDQN algorithm, as detailed in Algorithm~\ref{main_algo}, incorporates human interventions into the RL training process by integrating human actions at specified intervals, thereby allowing for real-time adjustments to the agent's policy based on human insights. Key parameters controlling the frequency and extent of human interventions include the intervention frequency ($h_{\text{freq}}$), the number of steps per intervention ($h_{\text{steps}}$), and the total limit on intervention steps ($H_{\text{limit}}$).

    \begin{algorithm}[ht]
    \caption{Interactive DDQN with Human-in-the-Loop}
    \begin{algorithmic}[1]  
        \fontsize{9pt}{9pt}\selectfont 
        \STATE Initialize Q-networks with weights $\theta_1$ and $\theta_2$
        \STATE Initialize target Q-networks with weights $\theta^-_1 = \theta_1$ and $\theta^-_2 = \theta_2$
        \STATE Initialize experience replay buffer $\mathcal{D}$, human prioritization $\alpha$
        \FOR{Episode = 1 to $M$}
            \STATE Initialize state $s$
            \WHILE{Not Done}
                \STATE Select action $a$ using an $\epsilon$-greedy policy
                \STATE Every $h$-steps up to $H_{\text{limit}}$ initiate human interaction
                \STATE Get human intervention signal $I_{s}$ and human action $a_h$
                \STATE Get agent's predicted action $a_{\scriptscriptstyle \text{DRL}}$
                \STATE Set action vector $\mathbf{a} = [a_{\scriptscriptstyle \text{DRL}}, a_h]$
                \STATE Execute action from: $a = [1 - I(s)]a_{\scriptscriptstyle \text{DRL}} + I(s) \times a_h$
                \STATE Observe reward $r$ and next state $s'$
                \STATE Store transition $d = (s, \mathbf{a}, r, s', \text{done})$ in online buffer $\mathcal{D}$
                \STATE Store transition $(d, I(s))$ in evaluative buffer $\mathcal{D}_{\text{store}}$
                \STATE Every $C$-steps do:
                    \STATE \quad Sample random mini-batch of transitions from $\mathcal{D}$
                    \STATE \quad Compute $Q_h$ and $Q_a$ for human and agent from $\mathbf{a}$
                    \STATE \quad Compute $Q_{\text{target}}$, $Q_{\text{combined}}$
                    \STATE \quad Compute TD errors and loss $L_i(\theta_i)$
                    \STATE \quad Update $\theta$ using gradient descent and perform soft update for $\theta^-$
            \ENDWHILE
        \ENDFOR
    \end{algorithmic}
    \label{main_algo}
    \end{algorithm}

\subsection{Evaluation Prediction Module (EPM)} \label{future_state_prediction}

We further introduce the \textit{Evaluation Prediction Module (EPM)}, a framework designed for the post-hoc evaluation of human interventions, performed offline after data collection rather than during real-time execution. The EPM framework, as detailed in Algorithm~\ref{predicting}, comprises two key components: a \textbf{Classifier Model} and a \textbf{Predictive Model}. The Classifier Model predicts the probability of a collision during a state transition \( s_t \to s_{t+1} \), while the Predictive Model forecasts the next state \( s_{t+1} \) and the corresponding reward \( r_{t+1} \), based on the current state \( s_t \) and a given action \( a \). Together, these models simulate what would have occurred if the human had not intervened by comparing the accumulated rewards of actions taken by the agent versus those taken by the human over a specified evaluation horizon. Given a state \( s \) and corresponding actions \( a_{\text{human}} \) and \( a_{\text{agent}} \):

    \begin{enumerate}
        
        \item \textbf{Classifier Model}, \( C(s, s'; \theta_c) \), predicts the probability of a transition from state \( s \) to \( s' \) resulting in a crash, formalized as: \( P(\text{crash} | s, s') = C(s, s'; \theta_c) \). 
    
        \item \textbf{Predictive Model}, \( O(s, a; \theta_o) \), predicts the next state \( s' \) and the associated reward \( r \) for a given current state \( s \) and action \( a \), expressed as:
        \( (s', r) = O(s, a; \theta_o). \)
    
    \end{enumerate}
    \begin{algorithm}[tb] 
    \caption{Evaluation Prediction Module}
    \begin{algorithmic}[1]
        \fontsize{9pt}{9pt}\selectfont
        \STATE Load models $O(s, a; \theta_o)$, $C(s, s'; \theta_c)$ and $Q(s;\theta_1)$
        \STATE Initialize steps for evaluation $N$, $\Sigma r_{\text{agent}} = 0, \Sigma r_{\text{human}} = 0$
        \FOR{transition $d_i = (s_i, [a_{\text{human}_i}, a_{\text{agent}_i}], r_i, s_i', \text{done}_i,I_i(s))$}
            \IF{$I_i(s) == 1$}
                \STATE $s_{\text{sim}}, a_{\text{sim}} = s_i, a_{{\text{agent}}_i}$
                \FOR{$j = i$ \textbf{to} $i+N$}            
                    \STATE $(s'_{\text{sim}}, r_{\text{sim}}) = O(s_{\text{sim}}, a_{\text{sim}}; \theta_o)$
                    \IF{not $C(s_{\text{sim}}, s'_{\text{sim}}; \theta_c)$} 
                        \STATE $\Sigma r_{\text{agent}} \mathrel{+}= r_{\text{sim}}$
                        \STATE $a_{\text{sim}} = Q(s_{\text{sim}}; \theta_1)$
                    \ELSE
                        \STATE $\Sigma r_{\text{agent}} = -1$
                        \STATE \textbf{break for}
                    \ENDIF      
                \ENDFOR
            \ELSE 
                \STATE $\Sigma r_{\text{human}} \mathrel{+}= r_i$
            \ENDIF
        \ENDFOR
        \STATE Compare $\Sigma r_{\text{agent}}$,  $\Sigma r_{\text{human}}$
    \end{algorithmic}
    \label{predicting}
    \end{algorithm}

\section{Experimental Setup}

\begin{figure}[h]
    \centering
    \includegraphics[width=1.0\linewidth]{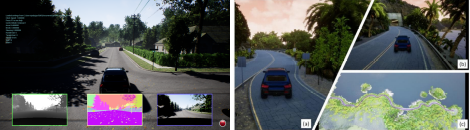}
    \caption{AirSim environments utilized for model training and evaluation. \textit{Left}: \textit{residential} training environment; \textit{right}: \textit{coastal} testing environment.} 
    \label{fig:airsim-environments}
\end{figure}

To evaluate the proposed iDDQN method, we utilized the AirSim simulation environment \cite{shah2018airsim}, which provides a high-fidelity simulation platform for autonomous driving tasks. Two distinct environments were designed for experimentation: the \textit{residential} environment for training and the \textit{coastal} environment for testing, as shown in Figure~\ref{fig:airsim-environments}. The training phase incorporated human interventions to guide policy learning, while the evaluation phase assessed the trained policies in the \textit{coastal} environment for measuring generalization performance.

Due to the high computational cost associated with human experiments, evaluations involving HITL interventions were conducted using two random seeds, while baseline models were evaluated with five random seeds. For the hyperparameters used in both the baseline models and proposed method, see Appendix \ref{AP: HYP}, Table \ref{tab:hyp-1}.

\subsection{Task and Episode Configuration} \label{task config}

The primary task of the autonomous vehicle agent was to navigate the environment while maintaining smooth driving behavior and avoiding collisions. Both environments included diverse challenges, such as varying lighting conditions and dynamic obstacles like animal crossings, to simulate realistic driving scenarios. The task was configured as follows:

\begin{itemize}
    \item \textbf{Speed and Brake Regulation:} The vehicle's speed was maintained between 32–40 km/h. Braking was excluded from the control set to simplify the learning process, focusing solely on steering control.
    
    \item \textbf{Driving Objective:} The agent was tasked with navigating the environment while minimizing collisions and ensuring smooth trajectory execution.
    
    \item \textbf{Action Space:} The agent's action space was a discrete 1D vector representing steering angle values ranging from \(-32^\circ\) to \(32^\circ\) in 2-degree increments, resulting in 33 possible actions. This granularity was chosen to balance precision and simplicity, as smaller differences in angles were empirically found to be imperceptible.
    
    \item \textbf{Episode Termination:} An episode concluded either when the cumulative reward exceeded 1000 (success), or the vehicle experienced a collision (failure).
\end{itemize}

\subsection{Human-in-the-Loop Intervention Process} \label{human intervention}

The HITL intervention process was designed to provide corrective feedback during the agent's learning phase. This process was implemented with the following components:

\begin{itemize}
    \item \textbf{Human Input Interface:} Experts interacted with a driving interface to deliver real-time corrective steering inputs. Prior to the experiments, participants were familiarized with the simulation environment to ensure their interventions were effective and consistent.
    
    \item \textbf{Intervention Dynamics:} Interventions were performed at regular intervals defined by the frequency parameter \(h_{\text{freq}}\) and lasted for \(h_{\text{steps}}\) steps, with a total cap of \(H_{\text{limit}}\). The corrective feedback was discretized to align with the agent’s action space for seamless integration.
\end{itemize}

\subsection{Reward Function} \label{total_reward}
\label{reward_func_sec}    
    Inspired by the works of \cite{salvador2019autonomous,spryn2018distributed}, we modify our reward components as the total reward \( r_{\text{total}} \) for the agent at each time step, being determined by the following:
    \begin{equation}
    r_{\text{total}} = 
    \begin{cases} 
    r_{\text{cr}} & \text{\textbf{if} car crashed} \\
    r_{\text{pos}} + r_{\text{sm}} & \text{otherwise} 
    \end{cases}
    \label{rewards}
    \end{equation}
    
    The constituting terms of \( r_{\text{total}} \) are:
    
    \begin{itemize}
        \item \textbf{Positional Reward (\( r_{\text{pos}} \))}: This reward incentivizes the vehicle to maintain an optimal position relative to the road's centerline. The hyperparameter \( \delta \) fine-tunes the exponential decay, while \( \beta \) sets a threshold for penalizing deviations. It is defined as:
       \begin{equation}
        r_{\text{pos}} = \min\left(e^{-\delta \times \left( \text{distance}^2 - \beta \right)}, 1\right)
    \end{equation}
     where $\text{distance}$ is calculated as the \textit{Euclidean} distance (excluding the z-axis) between the vehicle's current position \((x, y)\) and the nearest pre-recorded waypoint \((x_{wp}, y_{wp})\). 

          \item \textbf{Smoothness Penalty ($ r_{\text{sm}} $)}: This penalty encourages smoother steering transitions by computing the standard deviation of the agent's four latest steering actions. These decisions correspond to steering angles within the simulator's range of -0.8 to 0.8, effectively translating to actual angles between -32 to 32 degrees based on the agent's action selection from the Q-values. Given a buffer \( B \) containing the history of the last four executed actions \( a_{\text{sampled}} \), the penalty is formulated as:
    \begin{equation}
        r_{\text{sm}} = 
        \begin{cases} 
        0, & \text{if } |B| = 0 \\
        -\xi \times \sigma(b), & \text{otherwise} 
        \end{cases}
    \end{equation}

      \item \textbf{Crash Penalty (\( r_{\text{cr}} \))}: Negative reward is applied to strongly discourage crashes, penalized according to:
        \begin{equation}
        r_{\text{cr}} = -1
        \end{equation}
    \end{itemize}

    \label{eq:termination_condition}

\section{Results}
\subsection{Algorithm Comparison and Performance Evaluation}

The performance of iDDQN was evaluated in two stages. First, we analyzed the effect of varying the human weight factor \({\lambda_h}\) to assess the role of human guidance during training. Second, we identified the best-performing configuration and benchmarked it against baseline methods, including ``vanilla'' Clipped DDQN, DQfD, BC, and HG-DAgger.

\begin{figure}[ht]
    \centering
    \includegraphics[width=1\linewidth]{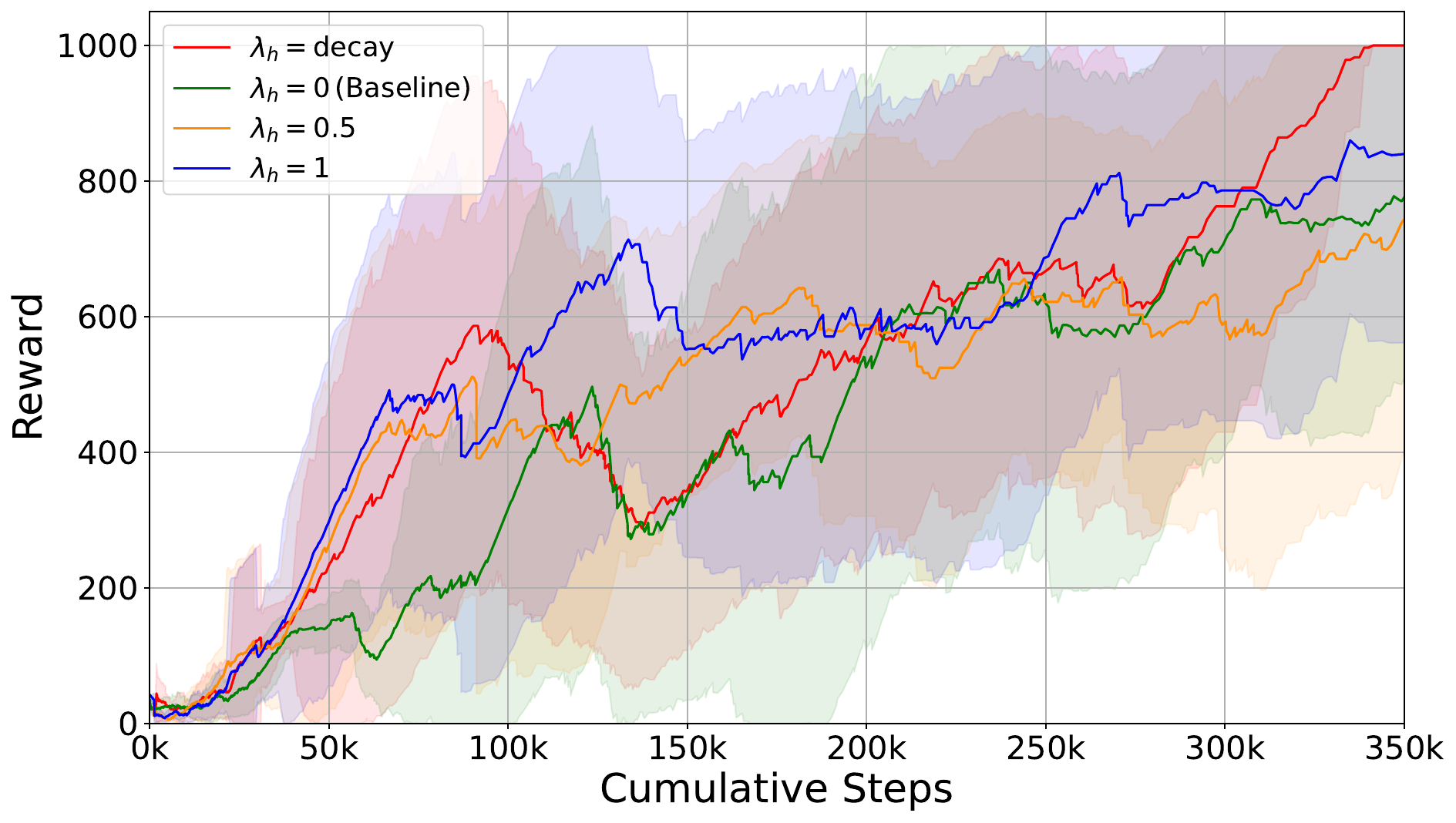}
    \caption{Performance comparison of iDDQN with varying \(\lambda_h\). The \(\lambda_h = \text{decay}\) configuration achieved the best results, balancing early human guidance and gradual autonomy.}
    \label{fig:results}
\end{figure}

In the first stage, we varied \(\lambda_h\) to explore human-agent collaboration. As shown in Figure~\ref{fig:results}, the \(\lambda_h = \text{decay}\) configuration achieved the best performance, leveraging early human guidance while transitioning to autonomous decision-making. Continuous human guidance \(\lambda_h = 1\) also improved training but was less efficient compared to the decay schedule. \(\lambda_h = 0.5\) demonstrated the lowest performance due to conflicts between human and agent inputs, while \(\lambda_h = 0\), representing basic RL without human input, performed moderately.

\begin{figure}[ht]
    \centering
    \includegraphics[width=1\linewidth]{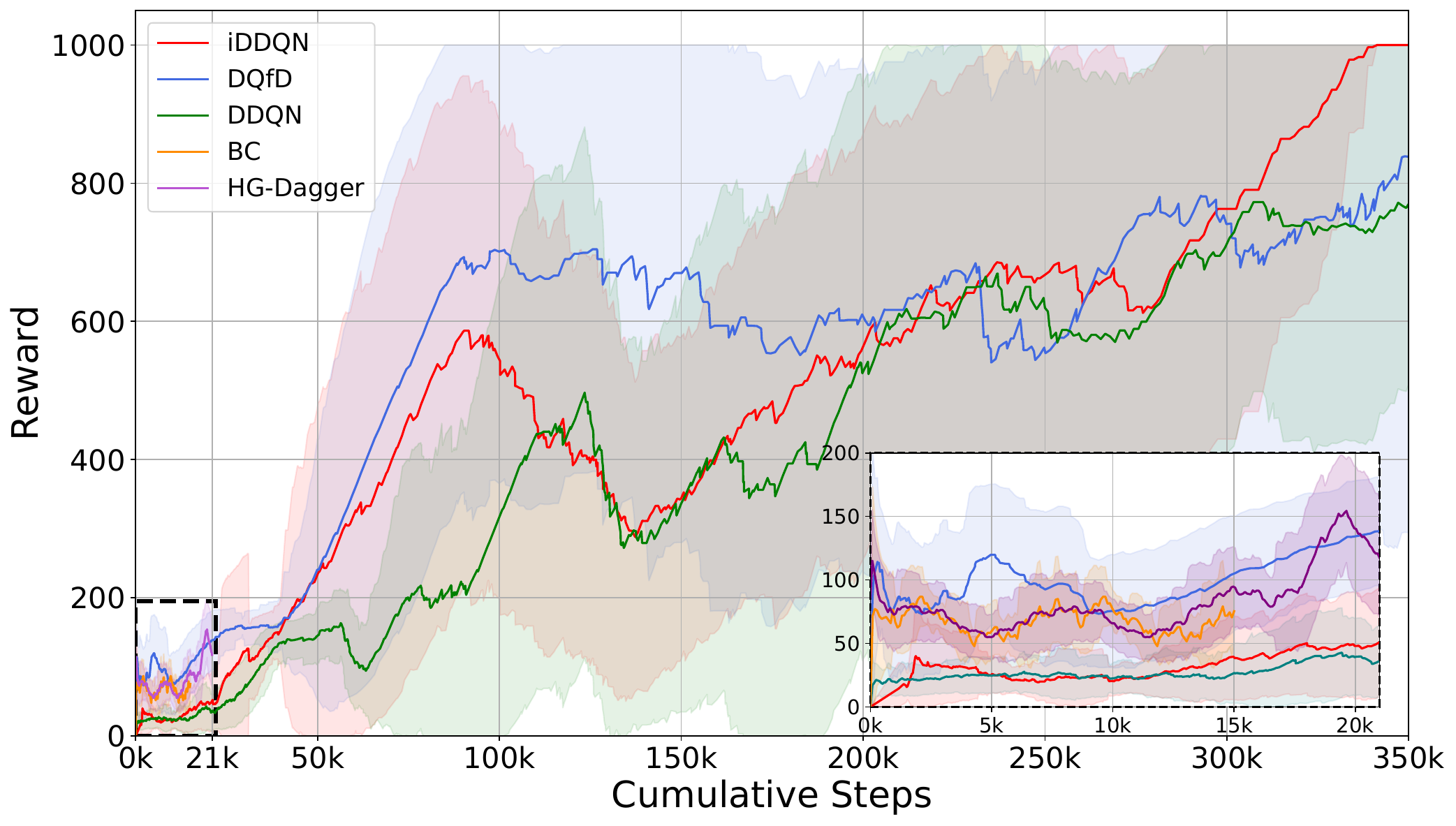}
    \caption{Comparison of episodic rewards during training for iDDQN and baseline methods. BC is trained for 15K steps (expert demonstrations) and HG-DAgger for 21K steps in total (after the 4th iteration). While BC and HG-DAgger struggled with distributional shift, DQfD leveraged pretraining effectively but was outperformed by iDDQN with the (\(\lambda_h = \text{decay}\)).}
    \label{fig:algo_comp}
\end{figure}

In the second stage, we benchmarked iDDQN (\(\lambda_h = \text{decay}\)) against HITL baselines. BC and HG-DAgger were trained with 15,000 initial expert demonstrations, with HG-DAgger iteratively adding 1,500 transitions until convergence. DQfD used expert demonstrations for pretraining, followed by reinforcement learning. Figure~\ref{fig:algo_comp} presents a comparison of episodic rewards, showing that iDDQN outperformed baseline methods, especially in the latter cumulative steps.   

\subsection{Generalization Across Unseen Environments}

To evaluate robustness, we tested the trained policies in an unseen \textit{coastal} environment (Figure~\ref{fig:airsim-environments}, right) over 100 episodes. Table~\ref{tab:policy_rollouts} summarizes the performance of iDDQN (\(\lambda_h\)-based configurations) compared to baseline methods. The results indicate that iDDQN (\(\lambda_h = \text{decay}\)) achieved the highest rewards and lowest variability, highlighting the far greater adaptability and superior performance in both environments.


\begin{table}[ht]
    \centering
\caption{Episodic rewards (mean ± standard deviation) for policy rollouts across Training and Testing Environments.}
    \label{tab:policy_rollouts}
    \begin{tabular}{lcc}
        \toprule
        \textbf{Method}         & \textbf{Training Env.} & \textbf{Testing Env. } \\
        \midrule
\(\boldsymbol{\lambda_h} = \text{decay}\) & \(\mathbf{858.90  \pm 130.31}\) & \(\mathbf{235.46 \pm 3.39}\) \\
        \( \lambda_h = 1.0 \)          & \( 812.24 \pm 165.19 \)              & \( 156.92 \pm 3.46 \)               \\
        \( \lambda_h = 0.5 \)          & \( 694.86  \pm 208.59 \)              & \( 82.17 \pm 3.44 \)                \\
        \( \lambda_h = 0 \) (DDQN)   & \( 762.00 \pm 201.24 \)              & \( 78.12 \pm 3.49 \)                \\
        DQfD                             & \( 784.37 \pm 180.45 \)              & \( 138.53 \pm 3.22 \)               \\
        HG-DAgger                        &  109.79 ± 46.23                               & 35.86 ± 2.61                                 \\
        BC         &  68.06 ± 25.23                                 & 19.11 ± 1.21                                \\
        \bottomrule
    \end{tabular}
    
\end{table}


\subsection{Alignment with Human Interventions}
To assess how closely human interventions align with agent-only behavior, we used the AirSim simulator to train and evaluate the proposed Evaluation Prediction Module (EPM), a post hoc analysis framework composed of a predictive model \( O(s, a; \theta_o) \) and a classifier \( C(s, s'; \theta_c) \). The predictive model forecasts next-state images and rewards using Structural Similarity Index Measure (SSIM)~\cite{wang2004ssim} and Mean Absolute Error (MAE), while the classifier estimates crash likelihood via binary cross-entropy loss.

\begin{figure*}[ht!]  
    \centering
    \includegraphics[width=1.0\linewidth]{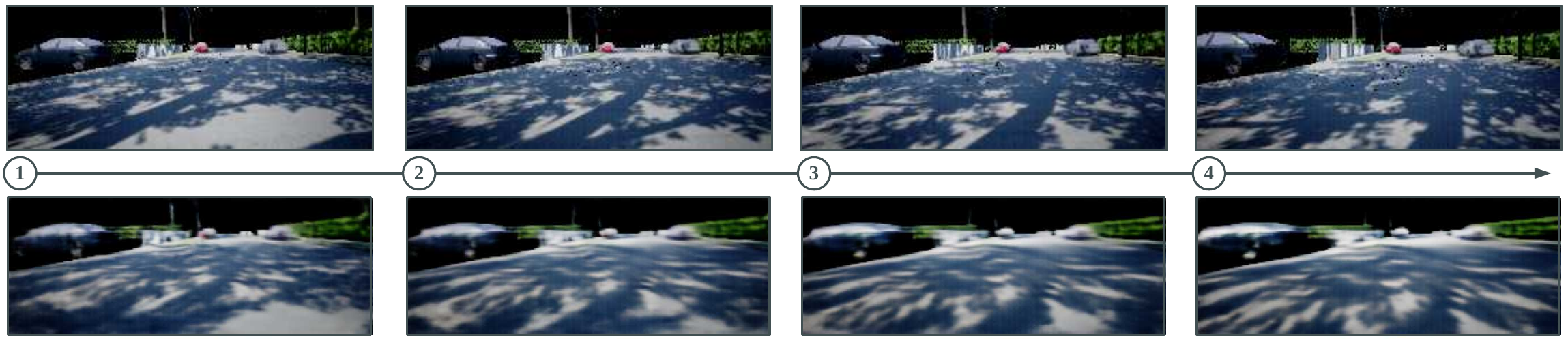}    
    \caption{The \textit{top row} depicts the actual trajectory executed when a human intervened. In contrast, the \textit{bottom row} shows the predicted trajectory generated by the EPM if the agent had acted autonomously without intervention. The cumulative reward achieved with human intervention (\(\sum r_{\text{actual}}=2.22\)) is higher than the reward predicted for the agent's decision (\(\sum r_{\text{agent}}=2.04\)), indicating that the EPM aligns with the human intervention. 
    }
    \label{fig:test_predictive}
\end{figure*}

The EPM is used offline to estimate what would have occurred had the agent acted without human input. It demonstrated strong alignment with human interventions, achieving a \textbf{94.2\% agreement rate} in cases where human actions led to higher cumulative rewards than the agent’s decisions, and only \textbf{5.8\% disagreement} in dynamic or ambiguous scenarios. Table~\ref{tab:epm_summary} summarizes the predictive model and classifier performance, and Figure~\ref{fig:test_predictive} visualizes the comparison between actual trajectories and EPM-predicted counterfactuals. Additional hyperparameters are listed in Table~\ref{tab:hyp-2}, Appendix~\ref{AP: HYP}.

\begin{table}[ht]
\centering

\caption{Performance Summary of the EPM Components}
\label{tab:epm_summary}
\begin{tabular}{llcc}
\toprule
\textbf{Model} & \textbf{Metric}                     & \textbf{Training Env.} & \textbf{Testing Env.} \\
\midrule
\multirow{2}{*}{\( C(s, s'; \theta_c) \)} 
    & Accuracy (\%)                & 93 ± 1.5                 & 83 ± 3.87              \\
    & F1 Score (\%)                & 92 ± 2.0                 & 80 ± 2.20              \\
\midrule
\multirow{2}{*}{\( O(s, a; \theta_o) \)} 
    & SSIM Loss                    & 0.26 ± 0.09             & 0.389 ± 0.10           \\
    & Reward MAE                   & 0.06 ± 0.01             & 0.12 ± 0.08            \\
\bottomrule
\end{tabular}
\end{table}

\section{Ablation Studies}

\begin{figure*}[htbp]
    \centering
    \begin{subfigure}[t]{0.32\textwidth}  
        \includegraphics[width=\linewidth]{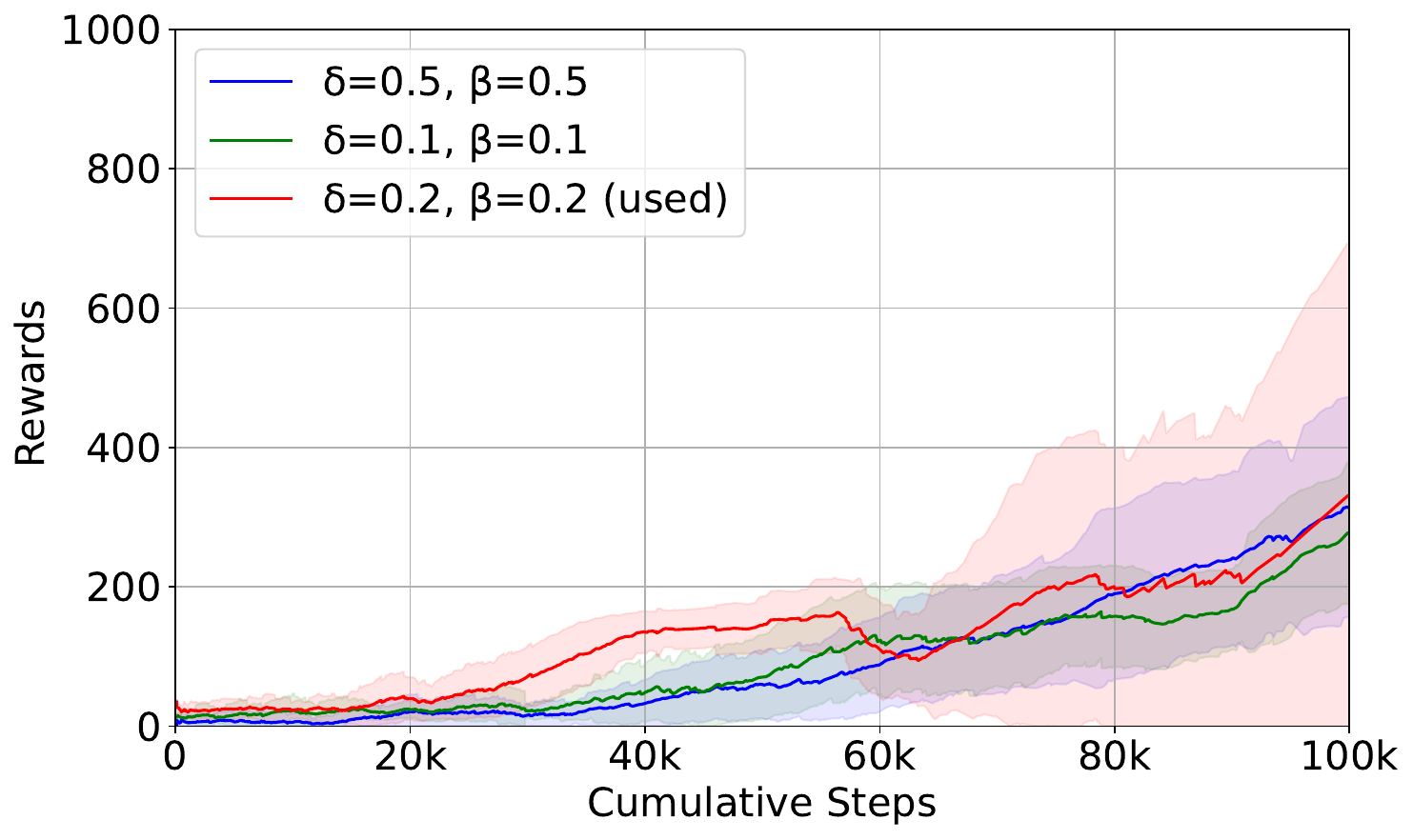}
        \caption{Hyperparameter exploration for \(r_{\text{pos}}\).}
        \label{fig:delta lambda}
    \end{subfigure}
    \hfill
    \begin{subfigure}[t]{0.32\textwidth}  
        \includegraphics[width=\linewidth]{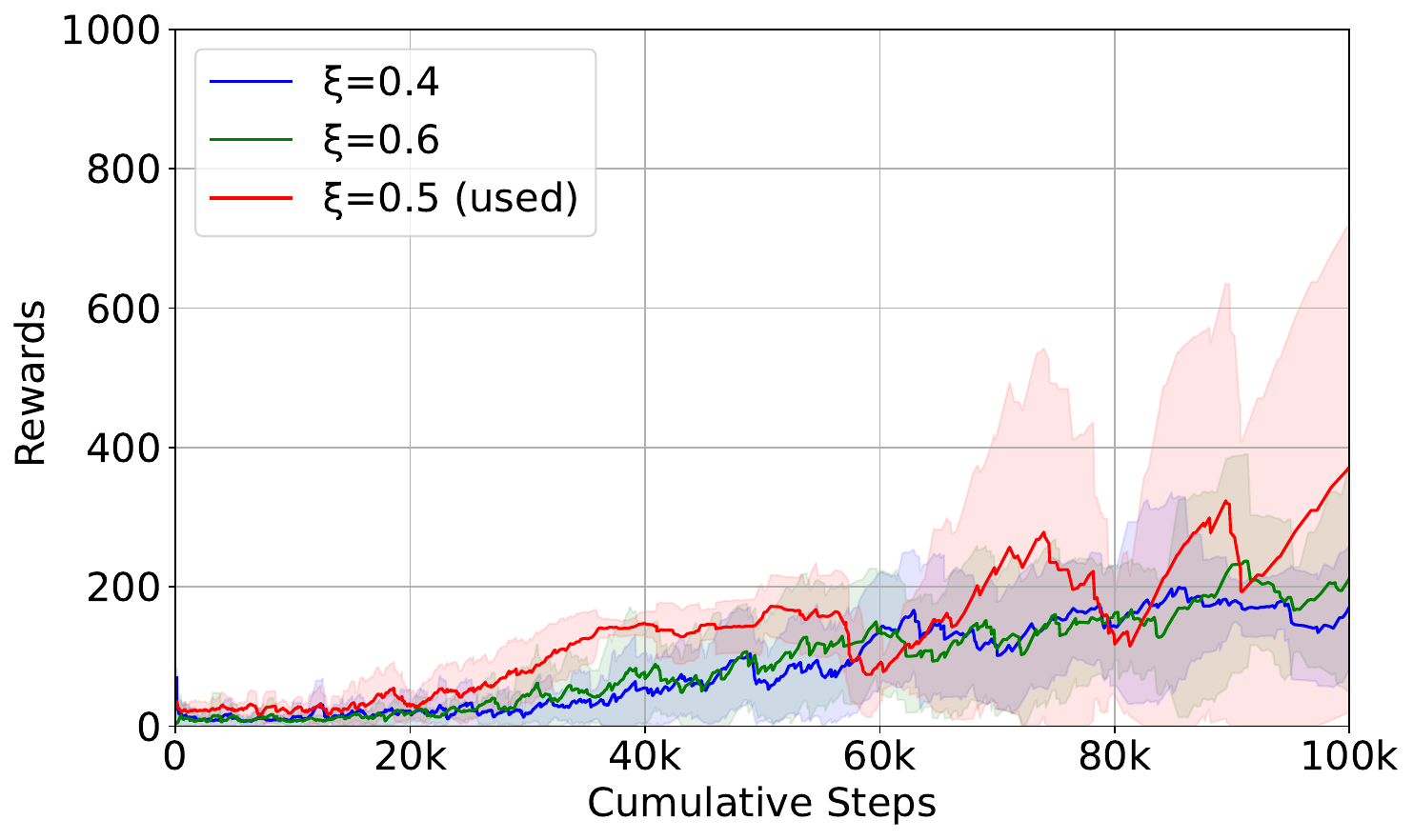}
        \caption{Hyperparameter exploration for \(r_{\text{sm}}\).}
        \label{fig:gamma}
    \end{subfigure}
    \hfill
    \begin{subfigure}[t]{0.32\textwidth}  
        \includegraphics[width=\linewidth]{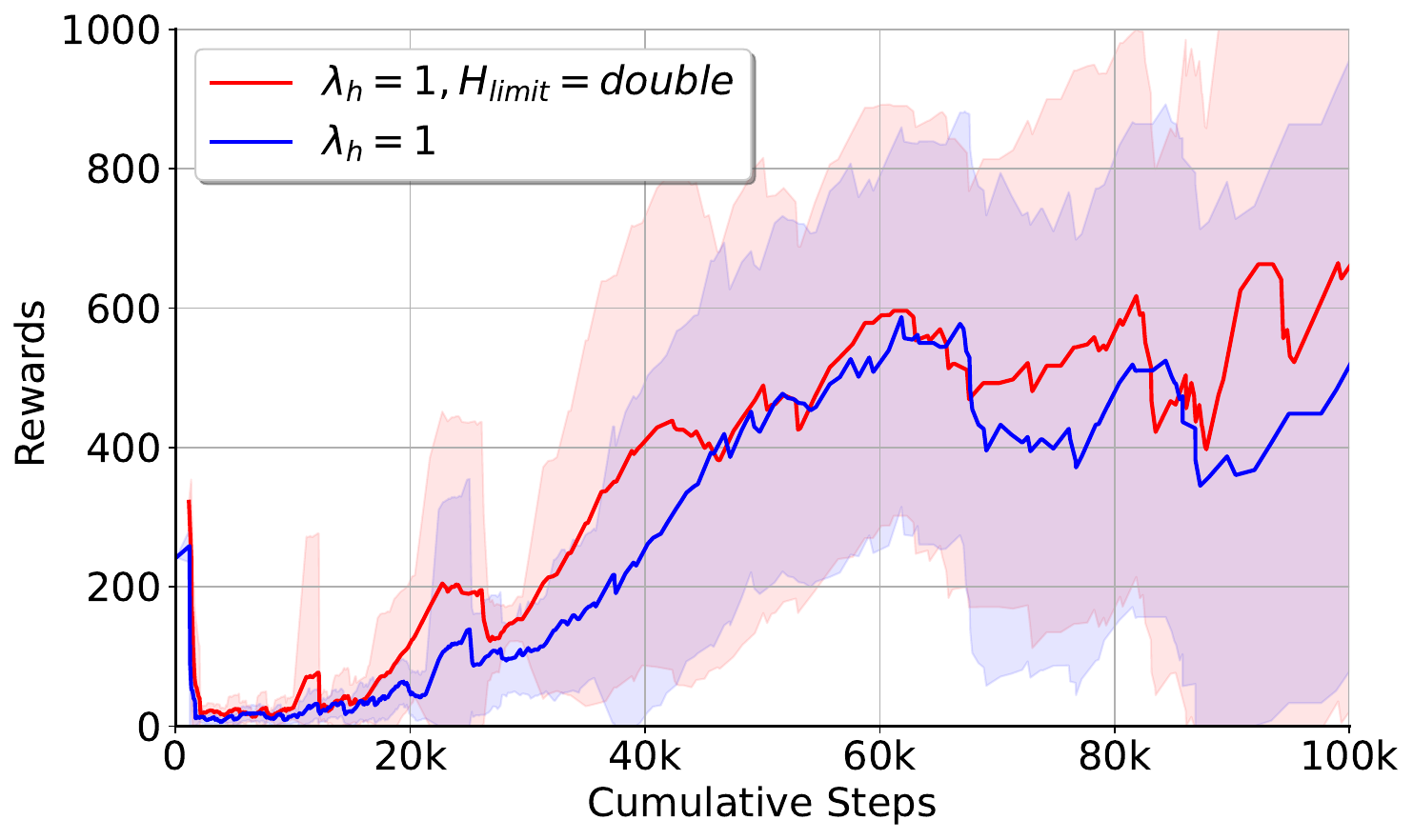}
        \caption{Impact of doubling human intervention frequency.}
        \label{fig:human_freq}
    \end{subfigure}
          \caption{Ablation studies for key hyperparameters. (\textbf{a}) Examines the effect of varying \(\delta\) and \(\beta\) on reward stabilization, showing a slight performance improvement at \(\delta=0.2\) and \(\beta=0.2\). (\textbf{b}) Analyzes \(\xi\), demonstrating better reward at \(\xi=0.5\) compared to other values. (\textbf{c}) Investigates the impact of doubling human intervention frequency for \(\lambda_h=1\), showing better early performance, but at the cost of doubling the intervention limit. This suggests that increasing human interventions does not necessarily result in proportional performance gains, highlighting the need for optimized intervention scheduling.}

    \label{fig:combined_ablation}
\end{figure*}

We conducted ablation studies to assess the effects of key hyperparameters on the performance of iDDQN. Specifically, we examined the impact of reward components (\(r_{\text{pos}}\) and \(r_{\text{sm}}\)) as defined in Section~\ref{total_reward} and the frequency of human interventions described in Section~\ref{pseud_oalgo}. In this ablation, we isolated and evaluated three aspects: (1) the influence of the positional reward decay factor \( \delta \) and threshold \( \beta \) on trajectory stability and overall rewards, (2) the effect of the smoothness penalty weighting \( \xi \) on agent control variability, and (3) the role of intervention frequency \( h_{\text{freq}} \) and the total intervention budget \( H_{\text{limit}} \) in shaping learning efficiency and policy robustness. The results of these ablation studies are presented in Figure~\ref{fig:combined_ablation}.

\section{Conclusion and Future Work}

This paper introduced iDDQN, a novel reinforcement learning approach that merges human interventions with a decaying influence parameter. Our results show that iDDQN outperforms baseline  methods (including BC, vanilla DRL, DQfD, HG-DAgger) in terms of training efficiency and adaptability. In addition, we proposed an Evaluation Prediction Module (EPM) introducing a systematic way to evaluate the impact of human actions, reinforcing the significance of human feedback in refining the agent’s policy.

Despite these promising results, certain limitations warrant further investigation. First, analyze rare cases where EPM and human interventions disagree. Second, the decay scheduling mechanism, while effective, remains static and may require contextual adjustments across different environments. Future work will explore dynamically adjusting the human weight factor, extending validation to real-world scenarios, and examining whether the gradual transition from human-guided actions to autonomous decisions enables the agent to adapt more effectively to novel environments, potentially improving real-world deployment success.



\section*{APPENDIX}
\subsection{Preliminaries}
\label{appendix_preliminaries}

Reinforcement Learning (RL) aims to learn an optimal policy \(\pi^*\) that maximizes the expected cumulative rewards. Given an observation state (\(s\)), action (\(a\)), reward (\(r\)), new state (\(s'\)), and  (\(done\)) symbolized as \( (s, a, r, s', done) \), which represent the environmental dynamics, the objective is defined as maximizing the sum of discounted future rewards:
\vspace{-0.1em}
\begin{equation}
R_t = \sum_{k=0}^{\infty} \gamma^k r_{t+k+1},
\end{equation}

where \(r_{t+k+1}\) represents the reward received at time \(t+k+1\), and \(\gamma\) is the discount factor, balancing immediate and future rewards. Hence, the optimal policy \(\pi^*\) prescribes the best action \(a\) in any given state \(s\) to maximize this cumulative reward.

\subsubsection*{Deep Q-Learning}

Deep Q-Learning utilizes deep neural networks to approximate the Q-function, \(Q(s, a; \theta)\), representing the expected reward for taking action \(a\) in state \(s\) following a policy \(\pi\). The goal is, therefore, to optimize the weights \(\theta\) of the neural network such that the Q-function reliably estimates the target Q-value (\(Q_{\text{target}}\)): 
\vspace{-0.1em}
\begin{equation}
Q_{\text{target}} = r + \gamma \max_{a'} Q(s', a'; \theta)
\end{equation}
where the term \( \max_{a'} Q(s', a'; \theta) \) selects the action \( a' \) in the next state \( s' \) that maximizes the Q-value.

\subsubsection*{Double Deep Q-Learning}

The Double Deep Q-Learning algorithm \cite{van2016deep} uses two neural networks to address the overestimation bias found in standard Q-Learning. These networks are denoted as \( Q(s, a; \theta) \) and the target \( Q(s, a; \theta^-) \). The key idea is to decouple the action selection from action evaluation, which mitigates the overestimation of action values and leads to more stable training:
\vspace{-0.1em}
\begin{equation}
Q_{\text{target}}= r + \gamma Q(s', \underset{a'}{\mathrm{argmax}}\ Q(s', a'; \theta); \theta^-)
\end{equation}

\subsubsection*{Dueling Architecture}

The Dueling Architecture, as proposed by \cite{wang2016dueling}, enhances Deep Reinforcement Learning by splitting the Q-function into two distinct streams: the state-value \( V(s; \theta) \) and the advantage-value \( A(s, a; \theta) \). Such structure allows the model to learn which states are valuable without the need to learn each action effect, for each state. The Q-value is instead computed by combining these two components via:
\vspace{-0.1em}
\begin{align}
    V(s; \theta) &= \mathbb{E}[Q(s, a; \theta)] \\
    A(s, a; \theta) &= Q(s, a; \theta) - V(s; \theta)
\end{align}

where the Q-value is calculated via:
\begin{equation}
    Q(s, a; \theta) = \quad V(s; \theta) + \left( A(s, a; \theta) - \frac{1}{|A|} \sum_{a'} A(s, a'; \theta) \right) 
\end{equation}

\subsubsection*{Clipped Double Deep Q-learning}

The Clipped Double Deep Q-learning approach \cite{fujimoto2018addressing} uses two separate main Q-networks, \( Q_1(s, a; \theta_1) \) and \( Q_2(s, a; \theta_1) \), to minimize overestimation bias by calculating the target Q-value. Thus, the target Q-value is subsequently computed as the minimum of the target Q-values predicted by each network:
\vspace{-0.1em}
\begin{equation} Q_\text{target} = r + \gamma \min_{i=1,2} Q_i(s', \text{argmax}_{a'} Q(s', a'; \theta); \theta^-_i)  (1-done)
\label{clipped}
\end{equation}

The Temporal Difference error (\( \text{TD}_{\text{error}} \)) in Clipped Double Q-learning is calculated using the primary (arbitrarily) network (e.g., \( Q_1 \)):
\vspace{-0.1em}
\begin{equation}
    \text{TD}_{\text{error}} = Q_{\text{target}} - Q_1(s, a; \theta_1) 
\end{equation}

\subsubsection*{Prioritized Experience Replay}

Prioritized Experience Replay (PER) enhances learning efficiency by focusing on transitions with higher Temporal Difference errors (\( \text{TD}_{\text{error}} \)), therefore prioritizing experiences that offer more significant learning opportunities:
\vspace{-0.1em}
\begin{equation}
    p_t = |\text{TD}_{\text{error}}| + \epsilon
\end{equation}
where \( p_t \) denotes the priority of transition \( t \), and \( \epsilon \) is a small constant ensuring all experiences have a non-zero chance of being sampled. Importance sampling weights \( w_t \) adjust for the bias introduced by this prioritized sampling:
\vspace{-0.1em}
\begin{equation}
    w_t = \left( \frac{1}{N \cdot P(t)} \right)^\beta
\end{equation}

where \( w_t \) represents the importance sampling weight for \( t \), \( N \) the replay buffer size, \( P(t) \) the sampling probability of \( t \), and \( \beta \) the bias correction parameter.

Incorporating the importance sampling weights in PER, the loss function is adjusted to account for the non-uniform sampling probabilities. It is defined as the weighted mean squared error of the Temporal Difference errors:
\vspace{-0.1em}
\begin{equation}
    L(\theta) = \mathbb{E}\left[ \sum_{t} w_t \cdot (\text{TD}_{\text{error,t}})^2 \right]
\end{equation}
 TD error for \( t \), which is computed according to:  
 \vspace{-0.1em}
  \begin{equation}
 \text{TD}_{\text{error,t}} =  Q_{{1},{\text{target,t}}} - Q(s_t, a_t; \theta).
 \end{equation}







\subsection{Hyperparameters} \label{AP: HYP}
The hyperparameter configurations are provided in Tables \ref{tab:hyp-1} and \ref{tab:hyp-2}. Table \ref{tab:hyp-1} lists key RL hyperparameters for baseline (vanilla RL) and iDDQN, including learning parameters and human intervention weighting. Table \ref{tab:hyp-2} compares shared hyperparameters between the classifier and predictive models. \textit{Geometric augmentations} include random rotations, zoom, and pixel noise applied to input images.

\begin{table}[ht]
\centering
\caption{Hyperparameter Configuration for RL Baselines and iDDQN}
\label{tab:hyp-1}  
\scriptsize  
\setlength{\tabcolsep}{3pt} 
\begin{tabular}{p{3.8cm}c}  
\toprule
\textbf{Parameter} & \textbf{Value} \\
\midrule
Learning Rate & 0.00025 \\
Optimizer & Adam \\
Gamma ($\gamma$) & 0.99 \\
Epsilon Init & 1 \\
Epsilon Decay & $1 \times 10^{-4}$ \\
Batch Size & 32 \\
Tau ($\tau$) & 0.0075 \\
Replay Buffer & 50000 \\
PER Alpha ($\alpha$) & 0.9 \\
Freq. Update Steps & 4 \\
Beta ($\beta$) & 0.4 \\
Action Space (Sim.) & [-0.8, 0.8] \\
Action Space (Real) & [$\pm32^\circ$] \\
Image Res. & $100 \times 256 \times 12$ \\
Conv Layers & 5 (16-256 filters) \\
Dense Layers & $256 \rightarrow [128 \times 2] \rightarrow 33$ \\
Augmentations & Elastic, Affine \\
Sensor Inputs & 6 (Yaw, Pitch, $V_x$, $V_y$, Speed, Steering) \\
Intervention Frequency ($h_{\text{freq}}$) & Every 20K steps \\
Total Interventions ($H_{\text{limit}}$) & 5 \\
Intervention Frequency ($h_{\text{freq}}$) & Every 20K steps \\
Total Interventions ($H_{\text{limit}}$) & 5 \\
Human Weight Decay ($\lambda_h$) & Linear (1 → 0 over 80K steps) \\
HG-DAgger (convergence) & 4 iterations \\
Evaluation Horizon  & 4 steps  \\
\bottomrule
\end{tabular}
\end{table}

\begin{table}[ht]
\centering
\caption{Common Hyperparameters for Predictive and Classifier Models}
\label{tab:hyp-2}  
\scriptsize  
\setlength{\tabcolsep}{3pt}  
\begin{tabular}{p{3.8cm}cc}  
\toprule
\textbf{Hyperparameter} & \textbf{Predictive} & \textbf{Classifier} \\
\midrule
Learning Rate & 0.0002 & 0.0005 \\
Optimizer & Nadam & Nadam \\
Loss Function & SSIM+MAE & CCE \\
Batch Size & 64 & 64 \\
Dropout Rate & 0.4 & 0.4 \\
Activation & Mish & Mish \\
Gaussian Noise (Input Layer) & 0.1 & 0.1 \\
Augmentations & Geometric, Noise & Geometric, Noise \\
\bottomrule
\end{tabular}
\end{table}

\section*{ACKNOWLEDGMENT}
This work was supported by the Wallenberg AI, Autonomous Systems and Software Program (WASP) and the NEST-project \texttt{\_\_main\_\_}.

\end{document}